# Building Resilience to Out-of-Distribution Visual Data via Input Optimization and Model Finetuning

Christopher J. Holder[1], Majid Khonji[2], Jorge Dias[2], and Muhammad Shafique[1]

[1]Division of Engineering, New York University Abu Dhabi, UAE
[2]Khalifa University Centre for Autonomous Robotic Systems, UAE

Corresponding author: Christopher J. Holder (e-mail: chris.holder@nyu.edu).

**ABSTRACT** A major challenge in machine learning is resilience to out-of-distribution data, that is data that exists outside of the distribution of a model's training data. Training is often performed using limited, carefully curated datasets and so when a model is deployed there is often a significant distribution shift as edge cases and anomalies not included in the training data are encountered. To address this, we propose the Input Optimisation Network, an image preprocessing model that learns to optimise input data for a specific target vision model. In this work we investigate several out-of-distribution scenarios in the context of semantic segmentation for autonomous vehicles, comparing an Input Optimisation based solution to existing approaches of finetuning the target model with augmented training data and an adversarially trained preprocessing model. We demonstrate that our approach can enable performance on such data comparable to that of a finetuned model, and subsequently that a combined approach, whereby an input optimization network is optimised to target a finetuned model, delivers superior performance to either method in isolation. Finally, we propose a joint optimisation approach, in which input optimization network and target model are trained simultaneously, which we demonstrate achieves significant further performance gains, particularly in challenging edge-case scenarios. We also demonstrate that our architecture can be reduced to a relatively compact size without a significant performance impact, potentially facilitating real time embedded applications.

## I. INTRODUCTION

This paper builds upon our prior work [1], in which we proposed the Input Optimisation Network (ION) to address the problem of poor lighting in urban driving scenarios. The original work is extended to include extreme weather scenarios as well as the distribution shift encountered between independently captured datasets, and introduces the concept of joint optimisation whereby ION training and model finetuning are conducted in parallel. Additional experiments are also included in which we validate our approach in a simple image classification task with the introduction of noisy input data.

Out of distribution (OOD) data poses a significant challenge in applied machine learning, as models can only be trained on data representative of a subset of the distribution of data that will be seen once deployed in a real-world setting. When a scenario is encountered that was not captured in a model's training data, performance can be significantly negatively impacted, leading to unreliable predictions. Examples of such data could include images affected by inclement weather or bad lighting, data captured in an environment not represented in the training data, or images capture by a camera with different properties to that used to capture training data.

To address the OOD problem, we propose an image-to-image convolutional neural network (CNN), that we term an Input Optimisation Network (ION), that learns to preprocess input data to more closely match the training distribution of a fixed pretrained target model by optimising for the performance of that model, effectively performing a kind of style transfer [2] where the target style is that which facilitates optimum performance of the target model. Unlike prior work on preprocessing OOD images that optimises for some hand-crafted metric, for example making images captured in rain look like images captured in clear weather [3], our approach automatically learns to prioritise image features that most benefit the performance of the target model.

The process for training the proposed ION is as follows: Firstly, a target model is trained to perform a specific task, for example image classification or segmentation, using only high-quality images. An ION is then trained to preprocess images that have been degraded in some way, for example with the addition of noise or simulated weather effects, with the goal of minimising the error within the outputs of the target model when these preprocessed images are used as its input. This error is then used to update the weights of the ION, while those of the target model remain fixed.

We demonstrate that via ION preprocessing, a target model trained only on high-quality images can achieve performance on degraded test data comparable to that of a model that has been fine-tuned on a similarly degraded training set. This opens up the possibility of successfully handling domain shift in scenarios where fine-tuning of a trained model may not be viable, for example when there is a risk of catastrophic forgetting.

We evaluate the proposed approach with an image classification task where inputs are degraded by image noise, and a semantic segmentation task where inputs are degraded to simulate inadequate lighting, rain, and fog, as well as a major domain shift scenario in which we introduce test images from a different dataset to that used in training our target model. In addition to comparing our approach to a model fine-tuning-based one, we also demonstrate that it significantly outperforms an adversarial training approach.

We then investigate the potential of a combined approach in which an ION is trained to target a finetuned model, i.e. degraded images are now included in the training distribution of the target model. Datasets intended to adequately represent wide varieties of real-world scenarios can involve multi-modal distributions with many difficult edge-cases that a model may not be robust to even when similar samples were included in its training data. In targeting such a finetuned model, an ION should learn to prioritise these edge cases and shift them closer to the distribution that the model performs well on.

Finally, we propose a joint optimisation approach, whereby ION and target model are trained simultaneously such that the target model optimises its weights for the images generated by the ION, rather than its weights remaining fixed. This enables both models to learn synergistically and converge on an optimal target distribution, rather than be constrained by the distribution of the original data the target model was trained for.

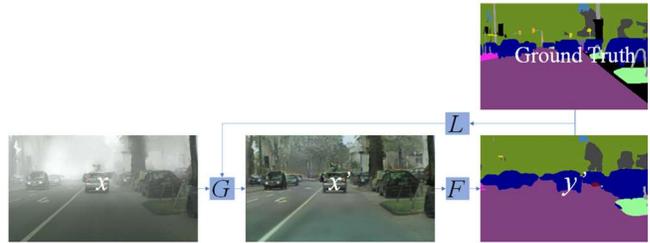

**Fig. 1.** The training pipeline of an ION optimising images for a semantic segmentation task. $G$ is the ION, $F$ is the trained target model, $x$ is the input image, $x' = G(x)$, $y' = F(x')$, and $L$ is the cross entropy loss between $y'$ and the ground truth. The objective of $G$ is to minimise $L$ while $F$ remains fixed.

**The main contributions of this paper are as follows:**
- We propose the concept of an Input Optimisation Network, a learned preprocessing model that facilitates performance on samples outside of a target model's training distribution comparable to that of a model whose training distribution included such images.
- We demonstrate that when dealing with complex, multi-modal datasets, combining ION preprocessing with finetuning of the target vision model enables performance better than either can achieve in isolation.
- We propose a joint optimization training scheme, whereby ION and target model are optimised simultaneously, demonstrating significantly improved performance, particularly in challenging edge-cases.

## II. RELATED WORK

Preprocessing of input data, specifically images, is a commonly used technique for improving the performance of machine learning models. In [4], simple statistical preprocessing techniques are compared using the performance of three basic convolutional neural networks, demonstrating significant improvement over non-processed input data. Correction of images that have been somehow degraded is a well-researched topic, motivated by both the need for data to remain in-distribution for vision models as well as human preferences for aesthetically pleasing images. Common causes of image degradation include noise [5][6], bad weather [3][7], bokeh or motion blur [8][9], and poor lighting [10]. Many recent approaches for correcting such images use generative adversarial neural networks [11] trained to remove specific phenomena from images, such as raindrops [3] or image blur [8]. Previously, hand-crafted approaches had been prevalent: In [12], two local contrast enhancement techniques are combined to improve the performance of face detection and recognition algorithms on poorly lit input images; In [13] a combination of colour, texture and shape features are used to detect rain drops which are then removed using image inpainting.

Another way to build resilience to OOD data is through training data augmentation or model finetuning, whereby a broader range of training data is used either to better represent that which will be encountered in the wild, or to improve a model's capability to generalise and therefore its robustness to OOD samples. One technique for shifting a model's expected distribution, albeit usually to a smaller, more specialized one, is Transfer Learning, or finetuning, whereby a model trained on a large dataset undergoes additional training on a smaller specialised target dataset [14], often with a subset of its weights fixed so that the parameter search space is smaller [15].

Another is to randomly apply one or more transformations to input data during training, either to build resilience to a specific phenomenon, such as in [11] where simulated fog is added to training images, or to increase the likelihood that the model learns to focus on salient features, and therefore to better generalise to unseen data, by removing potential confounding features. In [16], many transformations, including colour and geometric transforms, kernel filters and style transfer [2] are applied to training data and evaluated based on the ability of the resultant trained model to generalise to unseen test data. In [17] a technique called DeepAugment is proposed, in which images are passed through an image-to-image network whose weights and activations have been randomly distorted to create a wide array of image variations.

There is a significant body of work towards detection of OOD samples [18] [19] [20], the main motivation being that model outputs from such samples can subsequently be treated with lower confidence, however most of this work does not aim to improve robustness to such samples. There is a limited body of algorithmic approaches addressing the OOD problem: In [21], the problem of unsupervised domain adaptation is addressed through a combined reconstruction-classification network that uses a joint embedding to learn a useful representation of unlabeled data. In [22], the concept of Risk Extrapolation is proposed, in which a model aims to equalise performance across domains such that it learns to focus only on salient features.

There is, however, very little work towards the concept of a learned preprocessing function optimised for a specific computer vision model: In [23], the idea of Unadversarial Examples is proposed, in which an optimised image patch or object texture is learned that maximises the probability of correct detection or classification by a specific model.

In this work we build upon prior preprocessing-based approaches that aim to 'fix' degraded input samples, however rather than targeting specific image enhancements, such as rain drop removal or contrast enhancement, we propose an approach in which a preprocessing model learns to holistically optimise input samples for the performance of a specific target model.

## III. INPUT OPTIMISATION NETWORK

Our proposed Input Optimisation Network is an image-to-image preprocessing network that learns to generate optimised versions of input images prior to their being input to a specific target vision model. An ION, $G$, is trained to optimise its output for input to a separate target model, $F$, that has already been trained to convergence: Let us say that $F$ has already been optimised to minimise $L(y)$, where $L$ is some cost function and $F(x) = y$ for some input $x$. In training $G$, we aim to minimise $L(y')$ where $G(x) = x'$ and $F(x') = y'$. In the case of a semantic segmentation problem, $x$ is an RGB image, $y$ is the class probability map output by $F$ and $L(y)$ is the pixelwise cross entropy between output $y$ and a given ground truth segmentation. This pipeline is illustrated in Fig. 1. Backpropagation is used to compute the gradient of $L(y')$ with respect to both model's weights, which is then used to adjust the weights of $G$ while those of $F$ remain fixed.

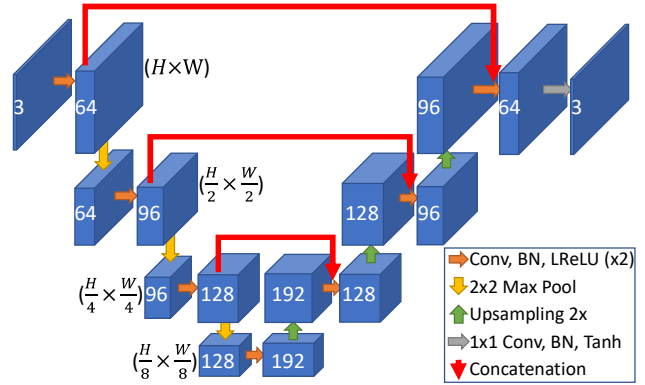

Fig. 2. The basic architecture for our ION preprocessing model is a symmetrical u-net style encoder-decoder network. Indices in white denote number of channels, brackets contain feature map size relative to initial input of $H \times W$.

We use a u-net-style [24] architecture with symmetrical encoder and decoder, each comprising $N$ blocks. An example of such an architecture where $N=4$ is shown in Fig. 2. Each encoder block is composed of two 3×3 convolutions, each followed by batch normalization and leaky rectified linear (ReLU) [25] operations, and a max pooling operation with a stride of 2. Input is of dimension $B \times C \times W \times H$, where $B$ is the number of samples in an input batch, $C$ is the number of channels or feature maps, and $W$ and $H$ are image width and height. In each subsequent block the number of feature maps, $C$, is increased while $W$ and $H$ are each decreased by a factor of 2. We set the number of feature maps of the first block, $C_1$, to 64, and increase $C$ in subsequent blocks using the Bulirsch sequence, i.e. 64, 96, 128, 192, 256…

The decoder mirrors the encoder, however pooling layers are replaced with bicubic upsampling operations and the output of each block is concatenated with the output of the corresponding encoder block via skip connection to maintain high-resolution details. The final layer of the model is a hyperbolic tangent function, which has been shown to aid learning in generative models [26], giving a three-channel output of the same dimensions as the input image with pixel values (-1, +1)

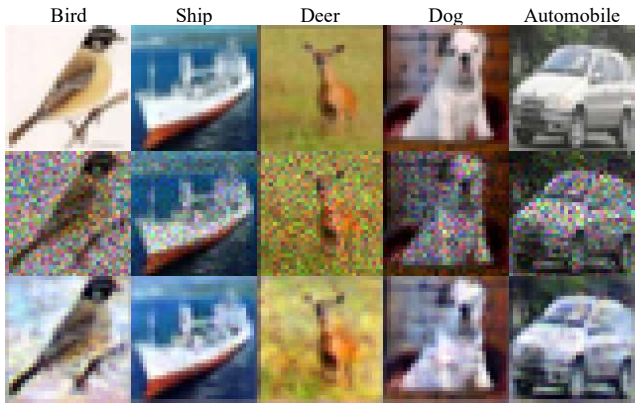

**Fig. 3.** Top: samples from the CIFAR-10 dataset; Middle: The same samples with noise added; Bottom: Noisy samples after processing by an ION trained to target a ResNet classification model. Note that the objective of the ION is not necessarily to remove noise, but to optimise an input image such that an accurate output is given by the target model.

In all our experiments, training continues until the ION converges using the Adam optimiser [27] with a fixed learning rate of $10^{-4}$ and regularisation via an L2 penalty of $10^{-6}$.

We validated this ION configuration during our experiments by comparing differently configured models. We trained ION models with leaky ReLU activation replaced by standard ReLU, with bicubic upsampling replaced by either nearest neighbour interpolation or learned transpose convolutional, with pooling operations replaced by strided convolutions and with the Adam optimiser replaced by stochastic gradient descent. None of these configurations were found to demonstrate any performance improvement. We evaluate models with different values selected for $N$ to investigate the trade-off between performance and compactness for potential real-time embedded scenarios.

## IV. EXPERIMENTAL SETUP

In this section, we describe the setup of each of the scenarios under which we evaluate the proposed approach. First, we validate the concept of the ION in a classification task with images degraded by Gaussian noise, using the CIFAR-10 [28] dataset and ResNet [29] and Inception [30] classification models. We then build on this with a semantic segmentation problem, evaluating the capabilities of our approach in the context of underexposed images in a road scene understanding scenario using the Cityscapes [31] dataset. Thirdly, we evaluate the ability of our approach to handle a complex multi-modal data distribution, wherein images impacted by rain and fog are combined with those from a totally different dataset, again in the context of semantic segmentation of road driving scenes. Finally, we describe our joint optimisation approach, in which ION and target model are trained simultaneously to achieve optimal results in a complex, multi-modal data scenario.

### A. CLASSIFICATION

We evaluate the capability of an ION to improve results when degraded images are used for a relatively simple classification task. For this we use the CIFAR-10 dataset [28], which contains 60,000 colour images of 32×32 pixels, divided into 10 classes of commonly recognizable objects. We create a degraded version of the CIFAR-10 dataset by adding randomly generated Gaussian noise to each image. The degraded training dataset is randomly generated at each training epoch, however the degraded test dataset is fixed to ensure consistency in our evaluation. Some examples from the test set are displayed in Fig. 3 along with the corresponding outputs from our trained ION model, in which it can be observed that in addition to removing noise, the ION appears to enhance the contrast and some colours in the images which may aid classifier performance.

We train two target classification models, ResNet 50 [29] and Inception V3 [30], on the original CIFAR-10 images (without noise added), and subsequently train two ION models, each targeting one of these classifiers. Due to the small input size of 32×32, the largest ION configuration that can be used is of N=4 blocks, as illustrated in Fig. 2.

During training, this model takes CIFAR-10 images, selected at random from either the original or noise-degraded set, as input and aims to produce images that minimize the negative log likelihood against the corresponding ground truth labels when passed through the target classification model. This error is used to update only the weights of the ION, while those in the target model remain fixed. We find that the best results are obtained from training with a learning rate of $10^{-4}$ and weight decay of $10^{-6}$ and using the Adam optimiser [27], with these parameters also used in all subsequent experiments.

We compare our approach to a fine-tuning approach, detailed in section 4.D., and evaluate a combined approach, wherein an ION is trained to optimise for a fine-tuned model. We also evaluate a joint optimisation approach, where ION and target model are trained simultaneously, detailed in section 4.E.

### B. UNDEREXPOSED IMAGERY IN SEMANTIC SEGMENTATION

#### 1) DATA

We train and evaluate our approach using two large-scale autonomous driving datasets: Cityscapes [31], which comprises 3475 RGB images (2975 for training, 500 for validation) with dimensions of 2048 × 1024 pixels along with semantic segmentation ground truth for 19 classes; and Audi Autonomous Driving Dataset (A2D2) [32], which comprises over 40,000 labelled images, from which we take a subset of 12,497 images (10,877 for training, 1620 for testing) with dimensions of 1920 × 1208 pixels. We relabel the A2D2 ground truth data using Cityscapes' labelling scheme to create a combined dataset of 13852 training images and 2120 test images (the Cityscapes validation set is used for testing as the ground truth labels for the test set have not been made publicly available).

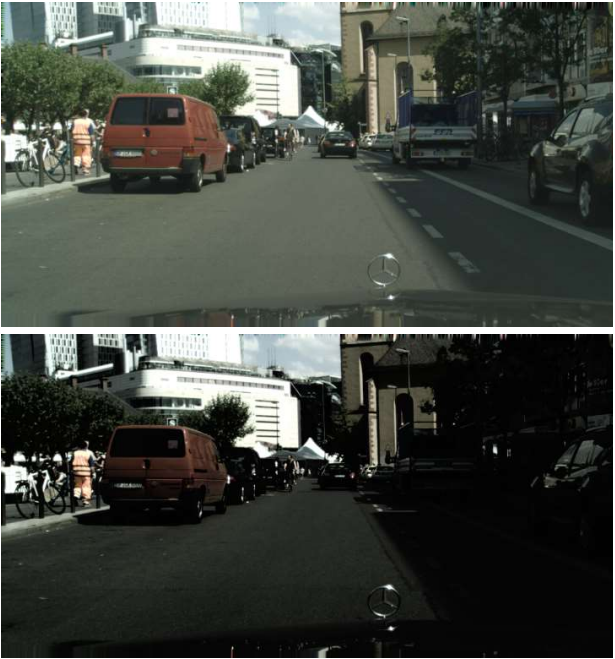

**Fig. 4**. An example image from the Cityscapes dataset [31] and a version degraded to simulate underexposure

We simulate the effects of underexposure by applying equation (1) to each pixel in an image:

$$V_2 = \begin{cases} \frac{V_1}{\theta_1/\theta_2}, & V_1 \leq \theta_1 \\ (1-\theta_2)\frac{V_1-\theta_1}{1-\theta_1} + \theta_2, & V_1 > \theta_1 \end{cases} \quad (1)$$

$V_1$ (0, 1) and $V_2$ (0, 1) are the brightness values (i.e. the V channel when the image is converted to HSV colour space) of the corresponding pixel in the original image and modified image respectively;

$\theta_1$ is a random threshold selected for each image such that $(\mu - \sigma) \leq \theta_1 \leq \mu$, where $\mu$ is the mean and $\sigma$ the standard deviation of all pixel values $V$ over the whole image;

$\theta_2$ is a second, lower threshold that determines by how much dark image regions are compressed. For our dataset we set $\theta_2 = \theta_1 \times 0.1$ such that pixels where $V_1 < \theta_1$ generate a corresponding $V_2 = V_1 \times 0.1$. Essentially this modifies the input image such that the dynamic range of pixel values below the threshold is compressed while that of pixel values above is expanded. An example image before and after application of equation (1) is shown in Fig. 4.

During model training, underexposed images are generated on-the-fly so that a different version of each image is seen during each epoch through the selection of different values for $\theta_1$, while our test set of simulated underexposed images is fixed so that different models can be compared consistently. Training images are cropped and resized to dimensions of 384 × 384 pixels, while test images are resized so that the largest dimension is 768 pixels with the original aspect ratio maintained. Training data is also augmented by random mirroring and cropping.

2) MODEL TRAINING

We use a u-net ION of size $N$=7 blocks as described in section III, optimising input images for a target semantic segmentation model using the DeepLab v3+ [33] architecture with a Mobilenet [34] backbone that has been trained using the Cityscapes dataset.

During training, input to our ION model are simulated underexposed versions of our combined training dataset of 13,852 images in batches of 4. Output are 3-channel RGB images that are subsequently used as input to the pre-trained semantic segmentation model. Error is computed as the mean pixelwise cross-entropy loss between the output of the segmentation model and the ground truth segmentation, and is backpropagated via the frozen segmentation model to update the weights of the generator network. Training continues until the model converges, which we found to happen after just 4 epochs.

3) GENERATIVE ADVERSARIAL NETWORK

We compare our approach with a Generative Adversarial Network (GAN) [11] using the same u-net-based architecture. When training a GAN, a separate discriminator network is trained simultaneously to differentiate between images from a desired distribution and those output by the generator network. The aim of the generator network is to create outputs indistinguishable from this desired distribution, in our case original images from the Cityscapes and A2D2 datasets.

Inputs are the same simulated underexposed images as used to train the ION model, and the GAN loss computed from the discriminator output is summed with the mean absolute error (L1 loss) between pixel values of the output image and the original, correctly-exposed version of that image to reduce the likelihood of an image's semantic content being altered. We found empirically that better results could be achieved by outputting a single channel image, rather than a 3-channel RGB image, which is then compared to only the V (value) channel of the target image in HSV colour space. This single channel is then combined with the H (hue) and S (saturation) channels of the input image to create the final image for segmentation. All other training parameters, including optimisation and batch size, are identical to those used in training the ION model.

*C. MULTIMODAL DOMAIN SHIFT*

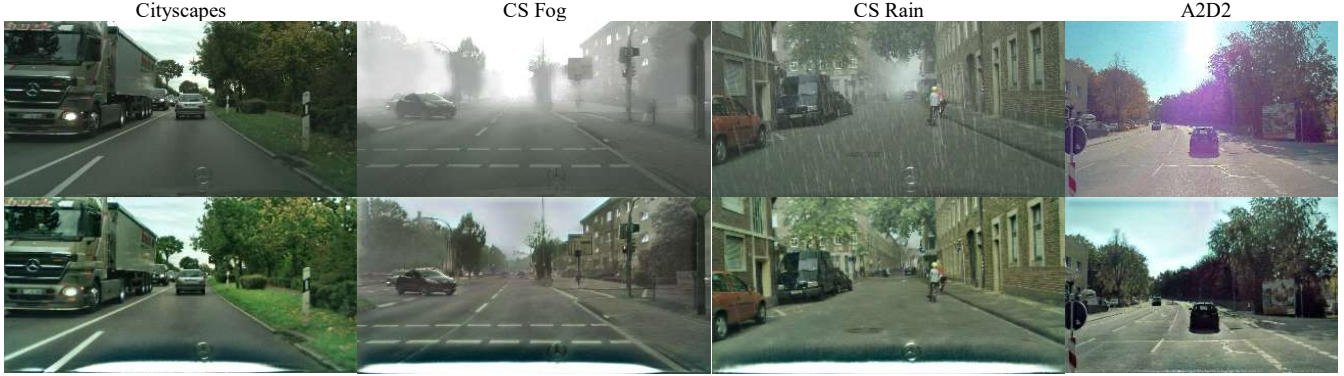

**Fig. 5.** Samples from our four segmentation datasets. top: original; bottom: processed by ION trained to target a segmentation model trained only on Cityscapes

In our third set of experiments, we evaluate our approach in a scenario where input samples can represent several different shifts away from the initial training distribution. We begin with a semantic segmentation model trained on the standard Cityscapes [31] dataset, and subsequently introduce images from the same dataset that have been degraded via simulated rain effects using the algorithm described in [35] (CS Rain), via simulated fog effects using the algorithm described in [7] (CS Fog), and images from a totally separate dataset that were captured by a different camera and in different environments (Audi Autonomous Driving Dataset (A2D2) [32]).

Our target segmentation model uses the Deeplab v3+ [33] architecture with a MobileNet [34] backbone that has been trained until convergence using the Cityscapes training dataset, an urban autonomous driving dataset comprising 3475 colour images of 2048×1024 pixels with corresponding pixelwise labels for 19 classes.

We evaluate this segmentation model using four datasets: The Cityscapes validation set (we use this as our test set due to ground truth labels not being available for the official test set), which comprises 500 images that can be assumed to share a similar distribution to the training set; the CS Fog and CS Rain validation sets, both of which degrade the original Cityscapes images with synthetic weather effects; and A2D2, another autonomous driving dataset comprising 12,497 images (of which we hold back 1620 as a test set) of 1920×1208 pixels with corresponding pixelwise labels, which we modify to align with the Cityscapes labelling scheme. It can be assumed that the images that make up CS Fog, CS Rain and A2D2 exist outside of the distribution of the segmentation model's training data.

We train an ION model to target this Cityscapes-trained Deeplab model using the training sets of the aforementioned four datasets. Output images are passed to the segmentation model and the pixelwise cross-entropy loss between the subsequent class probability map and ground truth segmentation is used to optimise the weights of the ION while those of the segmentation model remain fixed. During training, input images are randomly cropped to 384×384 pixels, and we use u-net-based ION configurations of size $N = 7$, $N = 5$ and $N = 3$ blocks to evaluate the performance impact of using a more compact model. During training we use an even mixture of samples from the Cityscapes, CS Fog, CS Rain and A2D2 training sets, with models observed to typically reach convergence after approximately 30 epochs of these combined datasets.

We evaluate these trained ION model using our four test datasets: each image is preprocessed by the ION and subsequently segmented by the target Deeplab model, with the class-wise intersection over union (IoU) of the resulting segmentation with the ground truth being our main evaluation metric. During testing, images are resized so that their width is 768 while maintaining their original aspect ratio.

Fig. 5 shows a sample from each of our test datasets and the corresponding output of our trained ION model. It can be seen that the in-distribution Cityscapes image remains relatively unmodified compared to the OOD images.

We compare our approach to a fine-tuning approach, detailed in section 4.D., and evaluate a combined approach, wherein an ION is trained to optimise for a fine-tuned model. We also evaluate a joint optimisation approach, where ION and target model are trained simultaneously, detailed in section 4.E. We chose not to evaluate a GAN-based approach for comparison in this scenario due to the poor performance observed in our prior experiments with underexposed imagery.

### D. FINETUNING

Many approaches for increasing resilience to OOD data involve data augmentation and model finetuning, whereby a model undergoes additional training with data that seeks either to better represent edge cases that may be encountered in the wild or to improve a model's ability to generalise to new data by broadening its training distribution such that it learns to focus on more salient features.

For comparison with our ION-based approach, we also finetune each of the target models using degraded or OOD data, essentially adding previously OOD data to their training

TABLE I
CLASSIFICATION RESULTS

| Model | Dataset | Baseline | ION | Finetuned | ION + Finetuned | Joint Opt Random | Joint Opt Pretrained | Joint Opt Finetuned |
|---|---|---|---|---|---|---|---|---|
| ResNet | Cifar-10 | **0.924** | 0.922 | 0.903 | 0.905 | 0.786 | 0.896 | 0.896 |
| ResNet | Noisy | 0.561 | **0.880** | 0.873 | 0.876 | 0.762 | 0.869 | 0.871 |
| Inception | Cifar-10 | **0.928** | 0.923 | 0.916 | 0.919 | 0.904 | 0.916 | 0.910 |
| Inception | Noisy | 0.5 | 0.866 | 0.879 | 0.889 | 0.877 | **0.893** | 0.885 |

Classification accuracy is shown for: baseline classification models trained only on original CIFAR-10 with no preprocessing at test time; baseline classification models with test data preprocessed by ION; finetuned classification model after additional training on both the original and noisy datasets; both techniques combined; Joint Optimised models where both models are randomly initialised (Joint Opt random); Joint Optimised models where the classification network has been pretrained using only the original Cifar-10 dataset (Joint Opt pretrained); and Joint Optimised models where the classification network has been pretrained with both original and noisy Cifar-10 datasets (Joint Opt finetuned). Bold indicates the best result for that row.

distribution. For our two classification networks, we take the models trained only using the original CIFAR-10 data and perform additional training using equal quantities of original and noise-degraded images. For our segmentation network, we take the model trained only using Cityscapes and perform additional training using an equal proportion of training samples from the Cityscapes, CS Fog, CS Rain and A2D2 datasets. We compare the performance of these finetuned models on the corresponding test sets to that of the original trained models on those same test sets after ION preprocessing. Our hypothesis is that the original models should perform as well on data outside of their training distribution that has been preprocessed by ION as the finetuned models whose training included similar data.

We also evaluate a combined approach, whereby ION models are trained using these finetuned models as targets, to demonstrate that when both ION and finetuning are used performance improves beyond what is achieved by either technique in isolation. In these experiments, the target model is finetuned as described above, and the ION training process is subsequently carried out using the resulting finetuned model, with weights frozen, as the target model.

### E. JOINT OPTIMISATION

The ION models discussed thus far have been trained to target a fixed model. We propose that performance can be improved by removing this constraint, allowing the target model to optimise for the data that its respective ION outputs. We implement this joint optimisation training scheme in an identical manner to the pipeline illustrated in Fig. 1 except that the loss function $L$ that is used to optimise the weights of ION $G$ is also used to optimise the weights of target model $F$ simultaneously. We evaluate this approach comparing three starting configurations of the target model: randomly initialised with no prior training; pretrained on just one dataset (CIFAR-10 or Cityscapes); or finetuned as described in 4.D. For classification, both the original and noise-degraded versions of CIFAR-10 are used, while for segmentation we use Cityscapes, CS Fog, CS Rain and A2D2. In all cases the ION model is randomly initialised with no prior training.

We train until we determine a model to have converged, when no further improvement in validation loss is observed.

With segmentation models this occurs after an average of approximately 40 epochs of the four combined datasets, compared to 30 epochs for the other ION models trained. To give an idea of training duration, one epoch takes approximately one hour using an Nvidia RTX 2080ti [36] GPU and a batch size of 4.

## V. RESULTS

### A. CLASSIFICATION

The metric we report for classification models is classification accuracy, i.e. the proportion of all samples in the test dataset for which the model assigns the correct class label. Results for both ResNet and Inception models, shown in Table I, are similar, with performance of baseline classifiers, trained only on the original CIFAR-10 images, falling by 40% to 50% when noisy test data is introduced. In all cases, ION preprocessing or finetuning results in a significant improvement in performance on the noisy test set and a slight reduction in performance on non-noisy test data. This drop in performance is smallest when the baseline classification model is combined with an ION, suggesting that preprocessing images to remove noise has less of a detrimental effect on non-noisy data than does the injection of noisy images into classification model training data.

In all cases, finetuning of the classification model combined with an ION trained to target this finetuned model gives better results than finetuning alone, however it does not surpass an ION working with the baseline model in most cases. Joint optimisation appears to have a consistently negative effect on ResNet performance, however this is not the case for Inception, where the joint optimised model with a classifier pretrained on just the clean CIFAR-10 data gave the best overall result on the noisy test set.

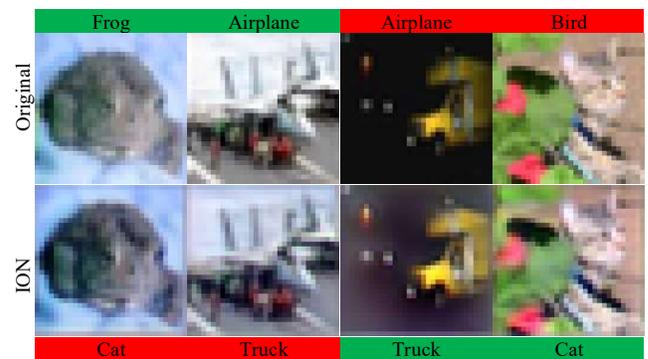

**Fig. 6.** Example images from CIFAR-10 without noise before (top) and after (bottom) being processed by an ION trained to target a ResNet classification model. In the left two columns, the original images were correctly classified while the ION images were not. In the right two columns, the ION images were correctly classified while the original images were not. Green identifies a correct prediction, red an incorrect prediction

TABLE II
CLASSIFICATION RESULTS

|  | Cityscapes | | | | Audi Autonomous Driving | | | |
|---|---|---|---|---|---|---|---|---|
|  | *Acc* | *Rec* | *Prec* | *IoU* | *Acc* | *Rec* | *Prec* | *IoU* |
| *Original* | 0.979 | 0.918 | 0.915 | 0.854 | 0.966 | 0.880 | 0.854 | 0.774 |
| *Dark* | 0.957 | 0.853 | 0.850 | 0.753 | 0.952 | 0.835 | 0.809 | 0.708 |
| *GAN* | 0.953 | 0.853 | 0.820 | 0.731 | 0.956 | 0.845 | 0.813 | 0.721 |
| *ION* | 0.973 | 0.895 | 0.898 | 0.824 | 0.973 | 0.883 | 0.881 | 0.805 |

Fig. 6 shows some examples of non-noisy CIFAR-10 images where ION preprocessing changed the output of the baseline ResNet classification model. In the left two cases a correct output became incorrect, and in the right two vice versa. We can observe that changes to the images are minor, with subtle alterations to contrast, particularly in dark regions, that in some cases help, but in others hinder, classification.

*B. UNDEREXPOSED IMAGERY IN SEMANTIC SEGMENTATION*

Outputs from our GAN and ION models trained to improve underexposed imagery are shown in Fig. 7, with corresponding segmentation results shown in Fig. 8. These examples show that the GAN learns to generate outputs that are visually very close to the original images, while the ION learns to generate oversaturated images with high contrast and strong edges, evidently features that the segmentation model benefits from. The image in the first column includes a cyclist on a dark street who becomes almost imperceptible in the underexposed image. Both GAN and ION models improve visibility of the cyclist, demonstrating the potential safety implications of this work. The third column features an image which appears to have already been slightly underexposed due to poor lighting, and while the GAN does a good job of recreating the original from the further underexposed image, the ION increases brightness, contrast and colour throughout the image, revealing a pedestrian in the distance (the blob segmented in red behind the third car) that the segmentation model was not able to detect in the original image.

We quantitively evaluate our approach using the Cityscapes and A2D2 test datasets, comprising 500 and 1620 images respectively. We create a single 'dark' version of each dataset using the method described in section IV.B.1 with a threshold chosen at random for each image, which we use as input for both ION and GAN models. The resulting outputs are then used as input to a trained DeepLab v3 semantic segmentation model, and we record the accuracy, recall, precision and intersection over union (IoU) of the resulting output against ground truth data. These 4 metrics are computed for each of the 19 classes present in the dataset, from which we calculate a weighted mean where each class

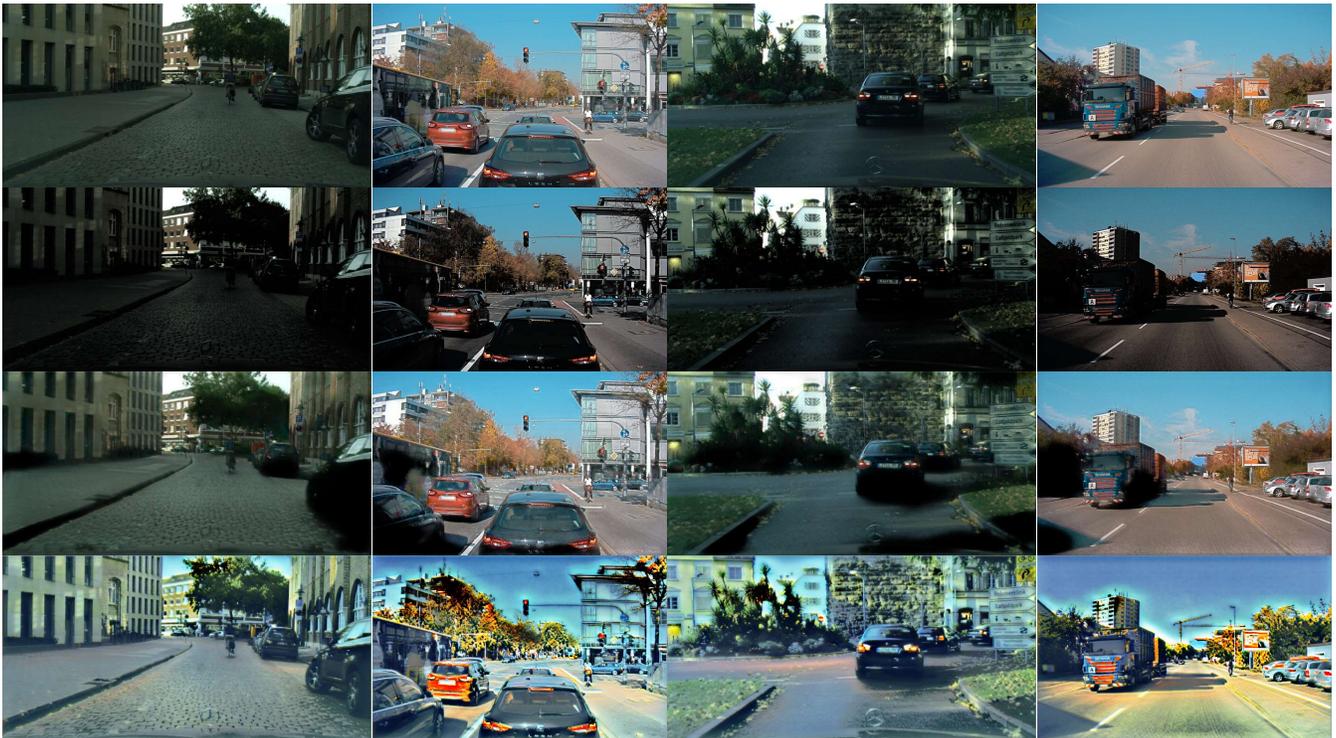

**Fig. 7.** Top row: original images from Cityscapes [31] and A2D2 [32] datasets. 2nd row: The same images modified to simulate underexposure. 3rd row: Output of our GAN model for corresponding underexposed input images. 4th row: Output of our ION model for corresponding underexposed images.

TABLE III
SEGMENTATION RESULTS

| Dataset | Baseline | ION | Finetuned | ION + Finetuned | Joint Opt Random | Joint Opt Pretrained | Joint Opt Finetuned |
|---|---|---|---|---|---|---|---|
| Cityscapes | 0.584 | 0.601 | 0.605 | 0.625 | 0.661 | **0.678** | 0.669 |
| CS Fog | 0.531 | 0.591 | 0.600 | 0.620 | 0.659 | **0.673** | 0.671 |
| CS Rain | 0.369 | 0.563 | 0.510 | 0.557 | 0.607 | **0.650** | 0.625 |
| A2D2 | 0.333 | 0.415 | 0.438 | 0.482 | 0.507 | **0.518** | 0.509 |

Mean class intersection over union for four semantic segmentation datasets. Baseline is a Deeplab model trained only on Cityscapes with no test time preprocessing. ION is a Deeplab model trained only on Cityscapes with images preprocessed by ION at test time. Finetuned is a Deeplab model fintuned on all four datasets with no test time preprocessing. ION + Finetuned combines both techniques. In Joint Opt both ION and Deeplab model are optimised simultaneously, with the deeplab model either randomly initialised, pretrained on just Cityscapes or finetuned on all four datasets before joint training begins. Bold indicates the best result for a given dataset.

is weighted by the proportion of pixels labelled as such throughout the dataset. For comparison, we also record segmentation results for the dark images and the original, unmodified images. These results are shown in Table II. It should be noted that one ION model and one GAN model, each trained on the combined Cityscapes/A2D2 training datasets, and one segmentation model, trained only on the Cityscapes training dataset, were used to generate both sets of results.

Our results for the Cityscapes dataset demonstrate the improvement our approach enables over the underexposed dark images, achieving an IoU close to that of the original images, while the GAN-generated images lead to worse performance than even the dark images.

Segmentation results are overall slightly worse for the A2D2 dataset than they are for Cityscapes, which is to be expected given that the segmentation model was trained using only the Cityscapes training dataset. In this case, both ION- and GAN-generated images improve performance over what was achieved using the dark images, with ION even improving performance over the original, correctly exposed images.

*C. MULTIMODAL DOMAIN SHIFT*
*1) SEMANTIC SEGMENTATION RESULTS*
Table III displays our semantic segmentation results, reporting mean class intersection over union (unlike in section V.B., classes are not weighted by frequency in computing IoU for this scenario). The performance of the baseline segmentation model, that was trained using only standard Cityscapes data, is expectedly poor on unseen datasets. Preprocessing images with an ION improves performance significantly, achieving results similar to those of the finetuned segmentation model that has been trained on all four datasets.

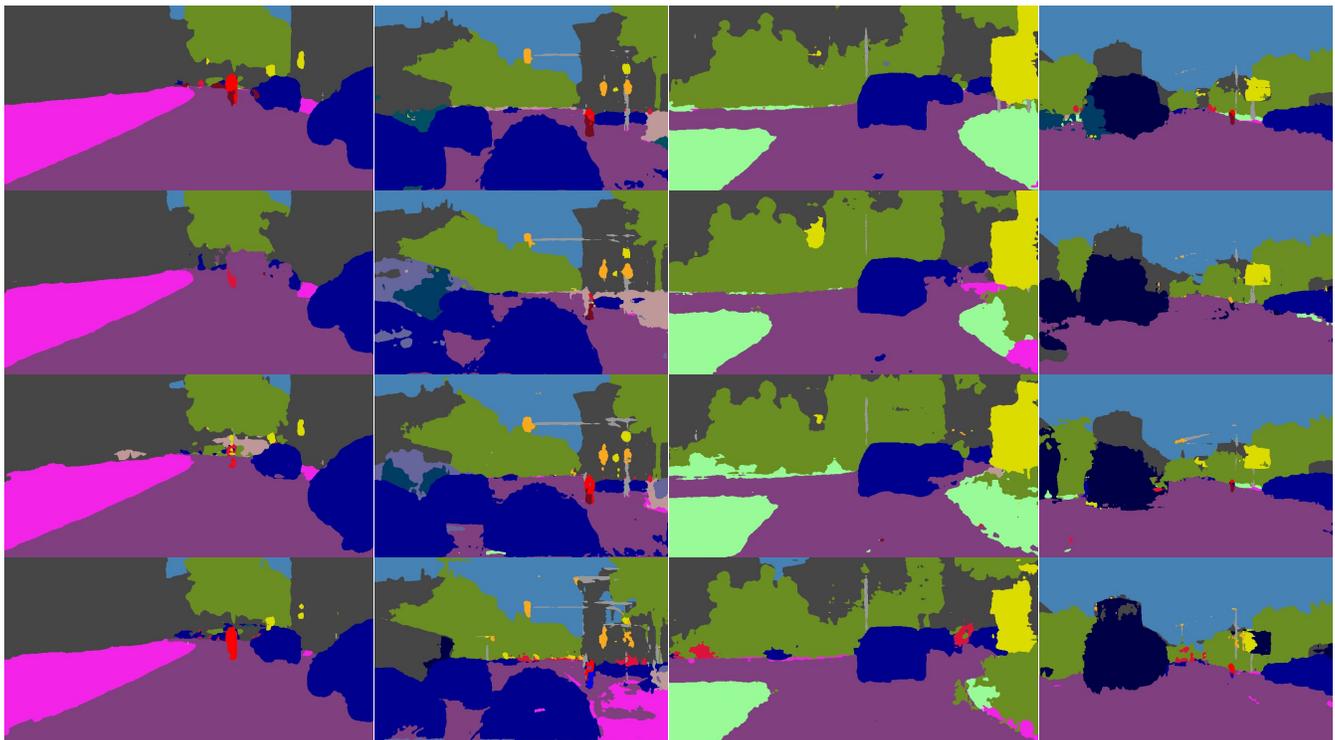

**Fig. 8.** Segmentation results from a Deeplab v3+ [33] model of the images in Fig. 7. Top row: original images. 2nd row: Images modified to simulate underexposure. 3rd row: Output from our GAN model. 4th row: Output from our ION model.

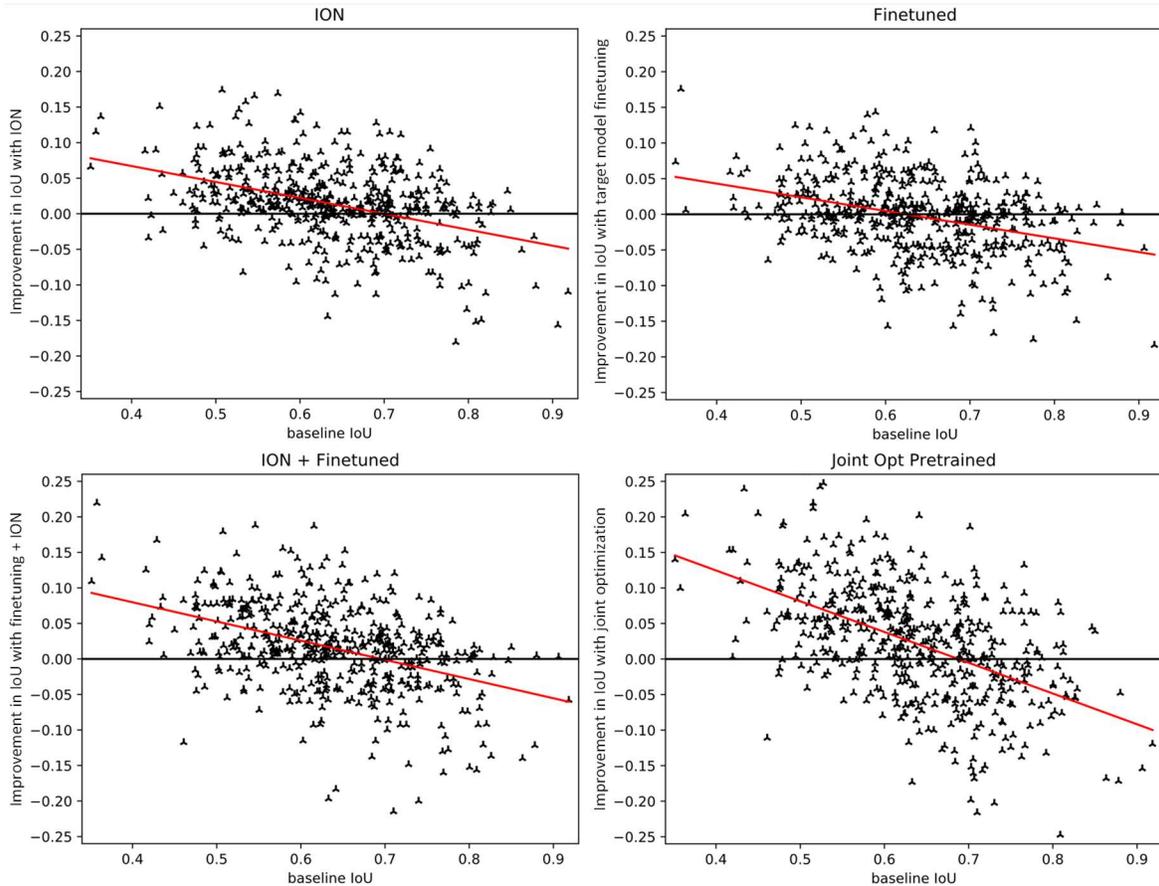

**Fig. 9.** Plots comparing the baseline intersection over union for each image in the Cityscapes dataset with the corresponding improvement that results from either ION preprocessing, model finetuning, combined finetuning and ION, or joint optimization using a pretrained segmentation model. Each point denotes a single image, with the x-axis representing the mean class IoU for that image from the baseline segmentation model, and the y-axis representing the amount by which the IoU changes when the corresponding technique is applied. The black line is y=0 i.e. points above represent an increase in IoU while points below represent a decrease. The red line is a linear polynomial fit to all points.

In both cases, we note an improvement in results from the standard Cityscapes dataset over those of the baseline segmentation model. In the finetuning case, this is likely due to the introduction of a broader set of training data improving the model's ability to generalize, boosting performance on edge-cases within the original dataset. In the ION case, this suggests a capability is learned for improving images that lie within, or close to, the distribution of the target model training data. For this reason, we investigate a combined approach whereby the segmentation model undergoes additional training using all four datasets and an ION model is subsequently trained to target this finetuned model. In this scenario, all of the data processed by the ION should be in-distribution for the target model, however this combined approach gives better results than either technique used in isolation.

Building further on this, we introduce our joint optimisation training scheme, whereby ION and target model are optimised simultaneously, bringing significant further performance improvement on all four datasets. We test this approach with three different levels of target network pretraining: random i.e. no pretraining; pretrained, where we use the baseline segmentation model trained only on Cityscapes as a starting point; and finetuned, where we use the finetuned segmentation model trained on all four datasets as a starting point. In all three cases the ION is randomly initialised. Our results show that better results are obtained when the target model has been pretrained, however additional finetuning appears to be detrimental. We speculate that this may be because the baseline model has been optimised for a smaller distribution of training data than the finetuned model, which has been optimised for a multimodal data distribution, giving the ION a more precise target distribution for its output images when the joint training begins.

Of particular interest are the performance gains seen for the original Cityscapes dataset: the baseline segmentation model had been trained to convergence on the Cityscapes training data, and yet all combinations of ION, finetuning and Joint Optimisation that are investigated bring about significantly better results on the corresponding test data. We believe this results from an ability to better deal with edge cases that exist within the test data that are not adequately represented in the training data, and so can still be considered a kind of weak OOD data. In Fig. 9 we test this idea by plotting the IoU recorded from the baseline segmentation model for each image in the Cityscapes test set against the subsequent improvement observed when each technique is applied. In all four cases, an inverse correlation exists between baseline IoU and the improvement subsequently observed, suggesting that the greatest benefit is seen with these edge-case samples that the original model performs poorly on. This correlation is stronger for ION preprocessing than for finetuning, increasing when both are combined and again when Joint Optimisation is introduced. In all cases there are several high-scoring samples that see a significant decrease in IoU, however the gains at the lower end are enough to raise the overall result. Future work could attempt to address this via an additional classification step that aims to predict whether or not an image should be preprocessed by ION or input directly to the target model.

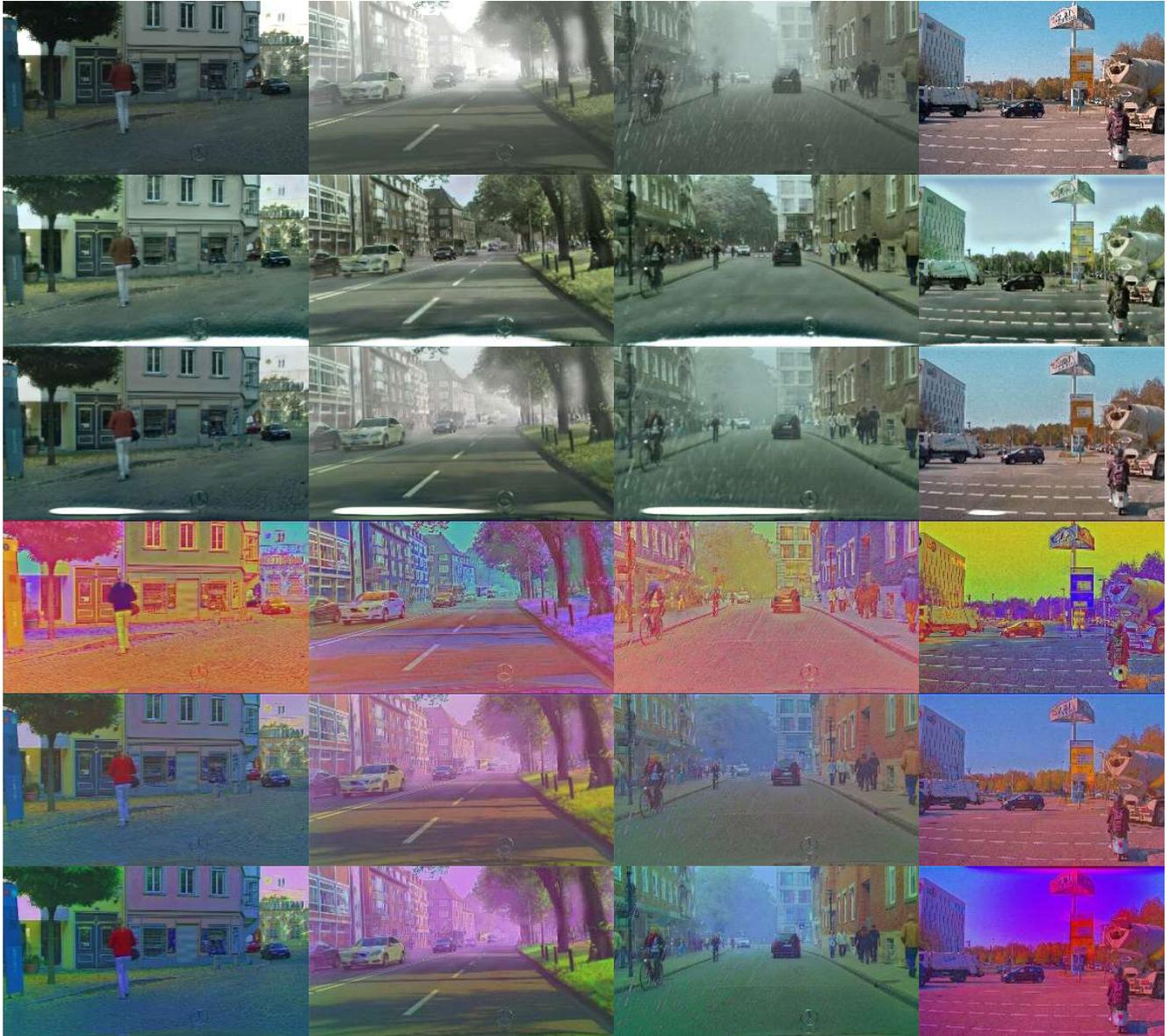

**Fig. 10.** Top: Original images from Cityscapes, CS Fog, CS Rain and A2D2 datasets respectively. 2nd row: Corresponding images output by an ION trained to target a segmentation model trained only on Cityscapes. 3rd row: output from an ION trained to target a segmentation model finetuned on all 4 datasets. 4th row: Joint optimised model with no pretraining. 5th row: Joint optimised model where the target model had been pretrained on Cityscapes. 6th row: Joint optimised model where the target model has been finetuned on all 4 datasets.

Fig. 10 shows output images of our different ION models from each dataset. We can see that the OOD images optimised for the baseline segmentation model have been modified to look more like in-distribution images: Rain and fog have been completely removed, while the A2D2 image appears to have been colour-corrected to appear more like a Cityscapes image, suggesting the ION learns to perform a kind of style transfer. When a target model that has been finetuned on all datasets is used, the changes are much subtler: rain and fog are reduced but not removed entirely, as the target model has already learned some resilience to their effects. When joint optimisation is introduced, images are

TABLE IV
COMPARISON OF RESULTS FROM DIFFERENT SIZED ION MODELS

| Dataset | Finetuned + ION | | | Joint Optimisation Random | | | Joint Optimisation Baseline | | | Joint Optimisation Finetuned | | |
|---|---|---|---|---|---|---|---|---|---|---|---|---|
| | 7-Block | 5-Block | 3-Block | 7-Block | 5-Block | 3-Block | 7-Block | 5-Block | 3-Block | 7-Block | 5-Block | 3-Block |
| Cityscapes | 0.625 | 0.624 | 0.615 | 0.661 | 0.637 | 0.582 | 0.678 | 0.683 | 0.670 | 0.669 | 0.681 | 0.678 |
| CS Fog | 0.620 | 0.618 | 0.610 | 0.659 | 0.637 | 0.573 | 0.673 | 0.678 | 0.668 | 0.671 | 0.681 | 0.683 |
| CS Rain | 0.557 | 0.559 | 0.559 | 0.607 | 0.599 | 0.532 | 0.650 | 0.642 | 0.621 | 0.625 | 0.615 | 0.650 |
| A2D2 | 0.482 | 0.488 | 0.477 | 0.507 | 0.496 | 0.450 | 0.518 | 0.518 | 0.518 | 0.509 | 0.533 | 0.528 |

Mean class intersection over union is reported for each test dataset. Light blue denotes a result from a smaller model that matches or surpasses that of the corresponding 7-block model, dark red a worse result.

altered much more drastically, as the target model is no longer fixed and can learn to adapt to ION output. The extreme colours seen are found to aid segmentation accuracy, despite not appearing realistic to a human eye. When the target model has been pretrained these effects are slightly reduced, presumably because the target model begins training already optimised for realistic-looking images and so the two models converge at an optimum closer to this than when the segmentation model is randomly initialised. Note that the best results were obtained with our baseline-pretrained joint optimisation model (5th row), which visually appears to have struck a balance between realism and extreme colours.

*2) MODEL SIZE*

All segmentation experiments thus far have utilised an ION architecture comprising an encoder and decoder each made up of 7 blocks as described in section III. We reperform some of these experiments using smaller 5- and 3- block models to assess the impact of model size on performance. Due to the increasing numbers of feature maps at deeper layers, decreasing the number of blocks in the encoder and decoder from 7 to 5 reduces the number of parameters in the model by a factor of approximately 4, and decreasing to 3 blocks reduces model size by the same factor again. However, as Table IV shows, our 5-block ION demonstrates similar – or in some cases a slight improvement in – performance, suggesting our original 7-block model is sub-optimally sized for the task at hand.

Results from the 3-block model are less consistent, with a marginal reduction in performance in most cases, however in the case of joint optimisation starting with a finetuned segmentation model, it is the 3-block ION that demonstrates the best overall performance, surpassing all other configurations on two of the four datasets used.

Using an Nvidia RTX 2080ti GPU [36] and processing images with dimensions of 768×384, the 7-block model runs at 11.9 fps, the 5-block model at 14.2 fps and the 3-block model at 16.5 fps.

## VI. CONCLUSIONS

In this work we have proposed a novel solution to the problems of degraded imagery and distribution shift in the input data of computer vision models. The proposed input optimisation network learns a preprocessing function optimised for a specific target vision model by attempting to process images such that they minimise the error in corresponding target model output. We evaluated the proposed approach in the contexts of image classification and semantic segmentation, in scenarios where images have been degraded by noise, underexposure, adverse weather and severe domain shift, and compared performance with adversarial and finetuning approaches.

We demonstrated that an ION is capable of learning to preprocess images that are outside of a target vision model's training distribution such that this model can achieve results comparable to that of a model that has been finetuned on similar images. We believe there are several notable use-cases where this technique could be further explored, such as training an ION to preprocess synthetic data for a target model trained on real data, enabling the creation of large-scale realistic-looking synthetic data sets. There may also be potential in computationally limited real-time applications such as robotics, where a compact neural network may struggle to generalise to complex scenarios and preprocessing could be a more efficient approach to increase robustness to edge-cases.

We have shown that the combination of model finetuning and input optimisation can improve performance across complex multi-modal datasets with many challenging edge-cases beyond that of either technique in isolation. In particular, we have shown that it is with the most difficult samples within these datasets that the most significant gains are observed.

Finally, we have proposed a novel joint optimisation training pipeline, where a preprocessing model and its target vision model are optimised simultaneously, demonstrating significant further performance increases when the constraint of a fixed target model is removed. We believe that in future this technique can not only be applied to increasing the performance of vision models, particularly in dealing with challenging edge-cases, but also contributing to model interpretability, through the analysis of the intermediate representations generated by an ION to aid identification of the salient features that a target model learns to rely upon.

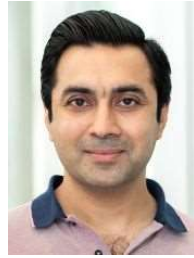

**Muhammad Shafique** (M'11 - SM'16) received the Ph.D. degree in computer science from the Karlsruhe Institute of Technology (KIT), Germany, in 2011. Afterwards, he established and led a highly recognized research group at KIT for several years as well as conducted impactful R&D activities in Pakistan and across the globe. In Oct.2016, he joined the Institute of Computer Engineering at the Faculty of Informatics, Technische Universitat Wien (TU Wien), ¨ Vienna, Austria as a Full Professor of Computer Architecture and Robust, Energy-Efficient Technologies. Since Sep.2020, he is with the Division of Engineering, New York University Abu Dhabi (NYU-AD), United Arab Emirates, and is a Global Network faculty at the NYU Tandon School of Engineering (NYU-NY), USA. His research interests are in brain-inspired computing, AI & machine learning hardware and system-level design, energy-efficient systems, robust computing, hardware security, emerging technologies, ML for EDA, FPGAs, MPSoCs, and embedded systems. His research has a special focus on cross-layer analysis, modeling, design, and optimization of computing and memory systems. The researched technologies and tools are deployed in application use cases from Internet-of-Things (IoT), smart Cyber-Physical Systems (CPS), and ICT for Development (ICT4D) domains. Dr. Shafique has given several Keynotes, Invited Talks, and Tutorials, as well as organized many special sessions at premier venues. He has served as the PC Chair, General Chair, Track Chair, and PC member for several prestigious IEEE and ACM conferences. Dr. Shafique holds one U.S. patent and has (co-)authored 6 Books, 10+ Book Chapters, 300+ papers in premier journals and conferences, and 50+ archive articles. He received the 2015 ACM/SIGDA Outstanding New Faculty Award, AI 2000 Chip Technology Most Influential Scholar Award in 2020, six gold medals, and several best paper awards and nominations at prestigious conferences like DAC, ICCAD, DATE, and CODES+ISSS.

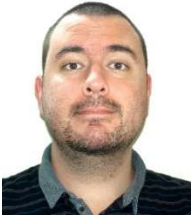

**Christopher J. Holder** received a BSc degree in Computer & Video Games from The University of Salford, UK, MSc degree in Computing from Cardiff University, UK, and PhD degree in Computer Science from Durham University, UK, focusing on the application of deep learning techniques to off-road autonomous driving. He has been a researcher at the Institute for Infocomm Research, Singapore, a postdoctoral researcher at Durham University, UK, Khalifa University, UAE and New York University Abu Dhabi, UAE, and is founder of Udara Limited, UK, an aerial imaging startup. His research focuses on the application of deep learning to visual problems, particularly localization, mapping, and scene understanding for robots and autonomous vehicles.

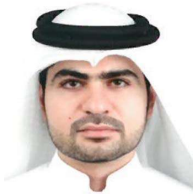

**Majid Khonji** received the M.Sc. degree in security, cryptology, and coding of information systems from the Ensimag, Grenoble Institute of Technology, France, and the Ph.D. degree in interdisciplinary engineering from the Masdar Institute, United Arab Emirates, in collaboration with the MIT in 2016. He is currently an Assistant Professor with the EECS Department, Khalifa University, United Arab Emirates, and a Research Affiliate with the MIT Computer Science and Artificial Intelligence Laboratory (CSAIL), USA. Previously, he was a Visiting Assistant Professor with the MIT CSAIL, a Senior Research and Development Technologist with Dubai Electricity and Water Authority (DEWA), and an Information Security Researcher with the Emirates Advanced Investment Group (EAIG).

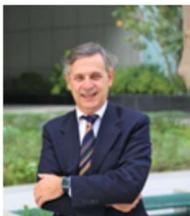

**Jorge Dias** has a Ph.D. on EE and Coordinates the Artificial Perception Group from the Institute of Systems and Robotics from the University of Coimbra, Portugal. He is Full Professor at Khalifa University, Abu Dhabi, UAE and Deputy Director from the Center of Autonomous Robotic Systems from Khalifa University. His expertise is in the area of Artificial Perception (Computer Vision and Robotic Vision) and has contributions on the field since 1984. He has been principal investigator and consortia coordinator from several research international projects, and coordinates the research group on Computer Vision and Artificial Perception from KUCARS. Jorge Dias published several articles in the area of Computer Vision and Robotics that include more than 300 publications in international journals and conference proceedings and recently published book on Probabilistic Robot Perception that addresses the use of statistical modeling and Artificial Intelligence for Perception, Planning and Decision in Robots. He was the Project Coordinator of two European Consortium for the Projects "Social Robot" and "GrowMeUP" that were developed to support the inclusivity and wellbeing for of the Elderly generation.